\documentclass[11pt,a4paper]{article}
\usepackage[hyperref]{naaclhlt2018}
\usepackage{times}
\usepackage{latexsym}

\usepackage{url}

\usepackage{natbib}
\usepackage{amsmath}
\usepackage{cleveref}

\usepackage{multirow}

\hypersetup{draft}

\aclfinalcopy 
\def\aclpaperid{20} 

\title{Incorporating Subword Information into Matrix \\ 
Factorization Word Embeddings}

\author{Alexandre Salle \quad Aline Villavicencio \\
  Institute of Informatics \\
  Universidade Federal do Rio Grande do Sul \\
  Porto Alegre, Brazil \\
  {\tt alex@alexsalle.com \quad avillavicencio@inf.ufrgs.br }}

\date{}

\begin{document}
\maketitle
\begin{abstract}
The positive effect of adding subword information to word embeddings has been demonstrated for predictive models. In this paper we investigate whether similar benefits can also be derived from incorporating subwords into counting models. We  evaluate the impact of different types of subwords (n-grams and unsupervised morphemes), with results confirming the importance of subword information in learning representations of rare and out-of-vocabulary words.
\footnote{This is a preprint of the paper that will be presented at the Second Workshop on Subword and Character LEvel Models in NLP (SCLeM) to be held at NAACL 2018.}
\end{abstract}

\section{Introduction}
\label{sec:intro}

Low dimensional word representations (embeddings) have become a key component in modern NLP systems for language modeling, parsing, sentiment classification, and many others. These embeddings are usually derived by employing the distributional hypothesis: that similar words appear in similar contexts \citep{Harris1954}. 

The models that perform the word embedding can be divided into two classes: predictive, which learn a target or context word distribution, and counting, which use a raw, weighted, or factored word-context co-occurrence matrix \citep{Baroni2014}. The most well-known predictive model, which has become eponymous with word embedding, is word2vec \citep{Mikolov2013}. Popular counting models include PPMI-SVD \citep{Levy2014}, GloVe \citep{Pennington2014}, and LexVec \citep{Salle2016}.

These models all learn word-level representations, which presents two main problems: 1) Learned information is not explicitly shared among the representations as each word has an independent vector. 2) There is no clear way to represent out-of-vocabulary (OOV) words.

fastText \citep{Bojanowski2017} addresses these issues in the Skip-gram word2vec model by representing a word by the sum of a unique vector and a set of shared character n-grams (from hereon simply referred to as n-grams) vectors. This addresses both issues above as learned information is shared through the n-gram vectors and from these OOV word representations can be constructed.

In this paper we propose incorporating subword information into counting models using a strategy similar to fastText. 

We use LexVec as the counting model as it generally outperforms PPMI-SVD and GloVe on intrinsic and extrinsic evaluations \citep{Salle2016a,cer2017semeval,Wohlgenannt2017UsingWE,Konkol2017GeographicalEO}, but the method proposed here should transfer to GloVe unchanged.

The LexVec objective is modified
\footnote{Our implementation of subword LexVec is available at https://github.com/alexandres/lexvec} 
such that a word's vector is the sum of all its subword vectors.

We compare 1) the use of n-gram subwords, like fastText, and 2) unsupervised morphemes identified using Morfessor \citep{virpioja2013} to learn whether more linguistically motivated subwords offer any advantage over simple n-grams. 

To evaluate the impact subword information has on in-vocabulary (IV) word representations, we run intrinsic evaluations consisting of word similarity and word analogy tasks. The incorporation of subword information results in similar gains (and losses) to that of fastText over Skip-gram. Whereas incorporating n-gram subwords tends to capture more syntactic information, unsupervised morphemes better preserve semantics while also improving syntactic results. Given that intrinsic performance can correlate poorly with performance on downstream tasks \citep{Tsvetkov2015EvaluationOW}, we also conduct evaluation using the VecEval suite of tasks \citep{Nayak2016}, in which 

all subword models, including fastText, show no significant improvement over word-level models.

We verify the model's ability to represent OOV words by quantitatively evaluating nearest-neighbors. Results show that, like fastText, both LexVec n-gram and (to a lesser degree) unsupervised morpheme models give coherent answers.

This paper discusses related word ($\S$\ref{sec:related}), introduces the subword LexVec model ($\S$\ref{sec:model}), describes experiments ($\S$\ref{sec:materials}), analyzes results ($\S$\ref{sec:results}), and concludes with ideas for future works ($\S$\ref{sec:conclusion}).

\section{Related Work}
\label{sec:related}
Word embeddings that leverage subword information were first introduced by \citet{schutze1993word} which represented a word of as the sum of four-gram vectors obtained running an SVD of a four-gram to four-gram co-occurrence matrix. Our model differs by learning the subword vectors and resulting representation jointly as weighted factorization of a word-context co-occurrence matrix is performed. 

There are many models that use character-level subword information to form word representations \citep{Ling2015FindingFI,Cao2016AJM,Kim2016CharacterAwareNL,Wieting2016CharagramEW,Verwimp2017CharacterWordLL}, as well as fastText (the model on which we base our work). Closely related are models that use morphological segmentation in learning word representations \citep{Luong2013,Botha2014CompositionalMF,qiu2014co,mitchell2015orthogonality,Cotterell2015MorphologicalW,Bhatia2016MorphologicalPF}. Our model also uses n-grams and morphological segmentation, but it performs explicit matrix factorization to learn subword and word representations, unlike these related models which mostly use neural networks.

Finally, \citet{Cotterell2016MorphologicalSA} and \citet{Vulic2017} 	\emph{retrofit} morphological information onto \emph{pre-trained} models. These differ from our work in that we incorporate morphological information at training time, and that only 
\citet{Cotterell2016MorphologicalSA} is able to generate embeddings for OOV words.

\section{Subword LexVec}
\label{sec:model}
The LexVec \cite{Salle2016a} model factorizes the PPMI-weighted word-context co-occurrence matrix using stochastic gradient descent.

\begin{equation}
PPMI_{wc} = max(0, \log \frac{M_{wc} \; M_{**}}{ M_{w*} \; M_{*c} })
\end{equation}
where $M$ is the word-context co-occurrence matrix constructed by sliding a window of fixed size centered over every \emph{target} word

$w$ in the 
\emph{subsampled} \citep{Mikolov2013}
training corpus and incrementing cell $M_{wc}$ for every \emph{context} word $c$ appearing within this window (forming a $(w,c)$ pair). LexVec adjusts the PPMI matrix using  \emph{context distribution smoothing} \citep{Levy2014}.

With the PPMI matrix calculated, the sliding window process is repeated and the following loss functions are minimized for every observed $(w,c)$ pair and target word $w$:
\begin{align}
\label{eq:lexvec2}
L_{wc} &= \frac{1}{2} (u_w^\top v_c - PPMI_{wc})^2 \\
\label{eq:lexvec3}
L_{w} &= \frac{1}{2} \sum\limits_{i=1}^k{\mathbf{E}_{c_i \sim P_n(c)} (u_w^\top v_{c_i} - PPMI_{wc_i})^2 }
\end{align}
where $u_w$ and $v_c$ are $d$-dimensional word and context vectors. The second loss function describes how, for each target word, $k$ \emph{negative samples} \citep{Mikolov2013} are drawn from the smoothed context unigram distribution.

Given a set of subwords $S_w$ for a word $w$, we follow fastText and replace $u_w$ in \cref{eq:lexvec2,eq:lexvec3} by $u'_w$ such that:
\begin{align}
\label{eq:sublex}
u'_w = \frac{1}{|S_w| + 1} (u_w + \sum_{s \in S_w} q_{hash(s)}) 
\end{align}
such that a word is the sum of its word vector and its $d$-dimensional subword vectors $q_x$. The number of possible subwords is very large so the function $hash(s)$\footnote{http://www.isthe.com/chongo/tech/comp/fnv/} hashes a subword to the interval $[1, buckets]$. For OOV words, 
\begin{align}
u'_w = \frac{1}{|S_w|} \sum_{s \in S_w} q_{hash(s)}
\end{align}

We compare two types of subwords: simple n-grams (like fastText) and unsupervised morphemes. For example, given the word ``cat'', we mark beginning and end with angled brackets and use all n-grams of length $3$ to $6$ as subwords, yielding $S_{\textnormal{cat}} = \{ \textnormal{$\langle$ ca, at$\rangle$, cat} \}$. Morfessor \citep{virpioja2013} is used to probabilistically segment words into morphemes. The Morfessor model is trained using raw text so it is entirely unsupervised. For the word ``subsequent'', we get $S_{\textnormal{subsequent}} = \{ \textnormal{$\langle$ sub, sequent$\rangle$} \}$.

\section{Materials}
\label{sec:materials}

Our experiments aim to measure if the incorporation of subword information into LexVec results in similar improvements as observed in moving from Skip-gram to fastText, and whether unsupervised morphemes offer any advantage over n-grams. For IV words, we perform intrinsic evaluation via word similarity and word analogy tasks, as well as downstream tasks. OOV word representation is tested through qualitative nearest-neighbor analysis.   

All models are trained using a 2015 dump of Wikipedia, lowercased and using only alphanumeric characters. Vocabulary is limited to words that appear at least $100$ times for a total of $303517$ words. Morfessor is trained on this vocabulary list.

We train the standard LexVec (LV), LexVec using n-grams (LV-N), and LexVec using unsupervised morphemes (LV-M) using the same hyper-parameters as \citet{Salle2016a} ($\textnormal{window} = 2$, $\textnormal{initial learning rate} = .025$, $\textnormal{subsampling} = 10^{-5}$, $\textnormal{negative samples} = 5$, $\textnormal{context distribution smoothing} = .75$, $\textnormal{positional contexts} = \textnormal{True}$).

Both Skip-gram (SG) and fastText (FT) are trained using the reference implementation\footnote{https://github.com/facebookresearch/fastText} of fastText with the hyper-parameters given by \citet{Bojanowski2017} ($\textnormal{window} = 5$, $\textnormal{initial learning rate} = .025$, $\textnormal{subsampling} = 10^{-4}$, $\textnormal{negative samples} = 5$). 

All five models are run for $5$ iterations over the training corpus and generate $300$ dimensional word representations. LV-N, LV-M, and FT use $2000000$ buckets when hashing subwords.

For word similarity evaluations, we use the WordSim-353 Similarity (WS-Sim) and Relatedness (WS-Rel)  \citep{Finkelstein2001} and SimLex-999 (SimLex) \citep{hill2015simlex} datasets, and the Rare Word (RW) \citep{Luong2013} dataset to verify if subword information improves rare word representation. Relationships are measured using the Google semantic (GSem) and syntactic (GSyn) analogies \citep{Mikolov2013} and the Microsoft syntactic analogies (MSR) dataset \citep{Mikolov2013b}.

We also evaluate all five models on downstream tasks from the VecEval suite \citep{Nayak2016}\footnote{https://github.com/NehaNayak/veceval}, using only the tasks for which training and evaluation data is freely available: chunking, sentiment and question classification, and natural language identification (NLI). The default settings   from  the suite are used, but we run only the \emph{fixed} settings, where the embeddings themselves are not tunable parameters of the models, forcing the system to use only the information already in the embeddings.

Finally, we use LV-N, LV-M, and FT to generate OOV word representations for the following words: 1) ``hellooo'': a greeting commonly used in instant messaging which emphasizes a syllable. 2) ``marvelicious'': a made-up word obtained by merging ``marvelous'' and ``delicious''. 3) ``louisana'': a misspelling of the proper name ``Louisiana''. 4) ``rereread'': recursive use of prefix ``re''. 5) ``tuzread'': made-up prefix ``tuz''. 
\section{Results}
\label{sec:results}

\begin{table}

\setlength\tabcolsep{5.4pt} 
  \begin{tabular}{|c|ccc|cc|}
\hline
Evaluation & LV & LV-N & LV-M & SG & FT \\
\hline
WS-Sim & .749 & .748 & .746 & \textbf{.783} & .778 \\
WS-Rel & .627 & .627 & .625 & \textbf{.683} & .672 \\
SimLex & .359 & \textbf{.374} & .366 & .371 & .367 \\
RW & .461 & \textbf{.522} & .479 & .481 & .500 \\
\hline
GSem & \textbf{80.7} & 73.8 & \textbf{80.7} & 78.9 & 77.0\\
GSyn & 62.8 & 68.6 & 63.8 & 68.2 & \textbf{71.1} \\
MSR & 49.6 & 55.0 & 53.8& 57.8 & \textbf{59.6} \\
\hline

Chunk & 90.4 & \textbf{90.6} & 90.5 & 90.4 & 90.4 \\
Sentiment & 77.0 & 77.0 & 77.6 & 75.3 & \emph{77.9} \\
Questions & \emph{87.4} & \emph{87.4} & 87.3 & 86.6 & 85.1 \\
NLI & 43.3 & 43.4 & 43.3 & 43.4 & \emph{43.8} \\
\hline
  \end{tabular}
\caption{Word similarity (\emph{Spearman's rho}), analogy (\% accuracy), and downstream task (\% accuracy) results. In downstream tasks, for the same model accuracy varies over different runs, so we report the mean over $20$ runs, in which the only significantly ($p < .05$ under a random permutation test) different result is in chunking.}
  \label{tab:intrinsic}

\end{table}

\begin{table*}[t]

  \centering
  \begin{tabular}{|c|c|l|}
\hline
Word & Model & 5 Nearest Neighbors \\
\hline
\multirow{3}{*}{``hellooo''} & LV-N & hellogoodbye, hello, helloworld, helloween, helluva \\
& LV-M & kitsos, finos, neros, nonono, theodoroi \\
& FT & hello, helloworld, hellogoodbye, helloween, joegazz \\
\hline
\multirow{3}{*}{``marvelicious''} & LV-N & delicious, marveled, marveling, licious, marvellous \\
& LV-M & marveling, marvelously, marveled, marvelled, loquacious \\
& FT & delicious, deliciously, marveling, licious, marvelman \\
\hline
\multirow{3}{*}{``louisana''} & LV-N & luisana, pisana, belisana, chiisana, rosana \\
& LV-M & louisy, louises, louison, louiseville, louisiade \\
& FT & luisana, louisa, belisana, anabella, rosana \\
\hline
\multirow{3}{*}{``rereread''} & LV-N & reread, rereading, read, writeread, rerecord \\
& LV-M & alread, carreer, whiteread, unremarked, oread \\
& FT & reread, rereading, read, reiterate, writeread \\
\hline
\multirow{3}{*}{``tuzread''} & LV-N & tuzi, tuz, tuzla, prizren, momchilgrad, studenica \\
& LV-M & tuzluca, paczk, goldsztajn, belzberg, yizkor \\
& FT & pazaryeri, tufanbeyli, yenipazar, leskovac, berovo \\

\hline
  \end{tabular}
\caption{We generate vectors for OOV using subword information and search for the nearest (cosine distance) words in the embedding space. The LV-M segmentation for each word is: $\{ \textnormal{$\langle$hell, o, o, o$\rangle$} \}$, $\{ \textnormal{$\langle$marvel, i, cious$\rangle$} \}$, $\{ \textnormal{$\langle$louis, ana$\rangle$} \}$, $\{ \textnormal{$\langle$re, re, read$\rangle$} \}$, $\{ \textnormal{$\langle$ tu, z, read$\rangle$} \}$. We omit the LV-N and FT n-grams as they are trivial and too numerous to list. } 
  \label{tab:oov}

\end{table*}

Results for IV evaluation are shown in \cref{tab:intrinsic}, and for OOV in \cref{tab:oov}.

Like in FT, the use of subword information in both LV-N and LV-M results in 1) better representation of rare words, as evidenced by the increase in RW correlation, and 2) significant improvement on the GSyn and MSR tasks, in evidence of subwords encoding information about a word's syntactic function (the suffix ``ly'', for example, suggests an adverb).   

There seems to a trade-off between capturing semantics and syntax as in both LV-N and FT there is an accompanying decrease on the GSem tasks in exchange for gains on the GSyn and MSR tasks. Morphological segmentation in LV-M appears to favor syntax less strongly than do simple n-grams.

On the downstream tasks, we only observe statistically significant ($p < .05$ under a random permutation test) improvement on the chunking task, and it is a very small gain. We attribute this to both regular and subword models having very similar quality on frequent IV word representation. Statistically, these are the words are that are most likely to appear in the downstream task instances, and so the superior representation of rare words 

has, due to their nature, little impact on overall accuracy. Because in all tasks OOV words are mapped to the ``$\langle$unk$\rangle$'' token, the subword models are not being used to the fullest, and in future work we will investigate whether generating representations for all words improves task performance.

In OOV representation (\cref{tab:oov}), LV-N and FT work almost identically, as is to be expected. Both find highly coherent neighbors for the words ``hellooo'', ``marvelicious'', and ``rereread''. Interestingly, the misspelling of ``louisana'' leads to coherent name-like neighbors, although none is the expected correct spelling ``louisiana''. All models stumble on the made-up prefix ``tuz''. A possible fix would be to down-weigh very rare subwords in the vector summation. LV-M is less robust than LV-N and FT on this task as it is highly sensitive to incorrect segmentation, exemplified in the ``hellooo'' example. 

Finally, we see that nearest-neighbors are a mixture of similarly pre/suffixed words. If these pre/suffixes are semantic, the neighbors are semantically related, else if syntactic they  have similar syntactic function. This suggests that it should be possible to get \emph{tunable} representations which are more driven by semantics or syntax by a \emph{weighted} summation of subword vectors, given we can identify whether a pre/suffix is semantic or syntactic in nature and weigh them accordingly. This might be possible without supervision using corpus statistics as syntactic subwords are likely to be more frequent, and so could be down-weighted for more semantic representations. This is something we will pursue in future work.

\section{Conclusion and Future Work}
\label{sec:conclusion}
In this paper, we incorporated subword information (simple n-grams and unsupervised morphemes) into the LexVec word embedding model and evaluated its impact on the resulting IV and OOV word vectors. Like fastText, subword LexVec learns better representations for rare words than its word-level counterpart. All models generated coherent representations for OOV words, with simple n-grams demonstrating more robustness than unsupervised morphemes. In future work, we will verify whether using OOV representations in downstream tasks improves performance. We will also explore the trade-off between semantics and syntax when subword information is used.

\bibliography{biblio}
\bibliographystyle{acl_natbib.bst}

\end{document}